\documentclass[journal]{IEEEtran}

\usepackage{cite}
\usepackage{amsmath,amssymb,amsfonts}
\usepackage{algorithmic}
\usepackage{graphicx}
\usepackage{textcomp}
\usepackage{url}
\usepackage{bm}
\usepackage{comment}

\newcommand{\argmax}{\operatornamewithlimits{argmax}}

\makeatletter
\AtBeginDocument{\DeclareMathVersion{bold}
\SetSymbolFont{operators}{bold}{T1}{ptm}{b}{n}
\SetSymbolFont{letters}{bold}{OML}{cmm}{b}{it}
\SetSymbolFont{symbols}{bold}{OMS}{cmsy}{b}{n}
\SetMathAlphabet{\mathrm}{bold}{T1}{ptm}{b}{n}
\SetMathAlphabet{\mathit}{bold}{T1}{ptm}{b}{it}
\SetMathAlphabet{\mathbf}{bold}{T1}{ptm}{b}{n}
\SetMathAlphabet{\mathtt}{bold}{OT1}{pcr}{b}{n}
\renewcommand\boldmath{\@nomath\boldmath\mathversion{bold}}}
\makeatother

\def\BibTeX{{\rm B\kern-.05em{\sc i\kern-.025em b}\kern-.08em
    T\kern-.1667em\lower.7ex\hbox{E}\kern-.125emX}}

\begin{document}

\title{Reinforcement Learning for Pollution Detection in a Randomized, Sparse and Nonstationary Environment with an Autonomous Underwater Vehicle}

\author{
    \IEEEauthorblockN{Sebastian Zieglmeier\IEEEauthorrefmark{1}, 
    Niklas Erdmann\IEEEauthorrefmark{1}, 
    and Narada D. Warakagoda\IEEEauthorrefmark{1}\IEEEauthorrefmark{2}}\\
    \IEEEauthorblockA{\IEEEauthorrefmark{1}Department of Technology Systems, University of Oslo, Kjeller, Norway}\\
    \IEEEauthorblockA{\IEEEauthorrefmark{2}Department of Defence Systems, Norwegian Defence Research Establishment, Kjeller, Norway}\\
}

\maketitle

\begin{abstract}
Reinforcement learning (RL) algorithms are designed to optimize problem-solving by learning actions that maximize rewards, a task that becomes particularly challenging in random and nonstationary environments. Even advanced RL algorithms are often limited in their ability to solve problems in these conditions. In applications such as searching for underwater pollution clouds with autonomous underwater vehicles (AUVs), RL algorithms must navigate reward sparse environments, where actions frequently result in a zero reward. This paper aims to address these challenges by revisiting and modifying classical RL approaches to efficiently operate in sparse, randomized, and nonstationary environments. We systematically study a large number of modifications, including hierarchical algorithm changes, multigoal learning, and the integration of a location memory as an external output filter to prevent state revisits. Our results demonstrate that a modified Monte Carlo-based approach significantly outperforms traditional Q-learning and two exhaustive search patterns, illustrating its potential in adapting RL to complex environments. These findings suggest that reinforcement learning approaches can be effectively adapted for use in random, nonstationary, and reward-sparse environments.
\end{abstract}


\section{Introduction}
\label{sec:introduction}
Classical reinforcement learning relies on reward functions to update a policy in a step-like manner through exploration of a task \cite{sutton2018reinforcement}. These methods depend heavily on consistent reward feedback to shape optimal solutions. However, in reward-sparse environments, where rewards are infrequent or zero, this approach becomes particularly challenging \cite{ocanaOverviewEnvironmentalFeatures2023}.

\let\thefootnote\relax\footnotetext{All code for this project is available at \url{https://github.com/SebAILab/HRL---pollution-cloud-detection-with-an-AUV.git}.}

One such challenging task arises when training autonomous underwater vehicles (AUVs) to search for water pollution clouds in the ocean. Water pollution, driven primarily by hydrocarbon emissions, poses an increasingly serious threat to marine ecosystems and local economies \cite{VINOTHKUMAR2020107932}. Early detection of leakages may prevent many of their worst environmental consequences. However, traditional detection methods are costly in terms of machinery and personnel \cite{marc2018}. As autonomous agents, AUVs are handled as a promising addition to the early detection efforts, but are limited by their limited battery capacity. Therefore, finding an efficient exploratory search approach is of great interest to mitigate the effects of environmental pollution. 

However, the given search task is far from trivial: A predefined search area can be supplied to an AUV, but the specific location or direction of the pollution cloud remains entirely unknown and random. The combination of reward sparsity, randomness, and a nonstationary environment causes the task to be extremely challenging. 
Further, pollution sensors on board the AUV have a limited reach, effectively only detecting a pollution cloud after it is reached.
In short, the problem is more akin to finding an optimal search pattern over an area in order to rapidly identify the source of pollution (target). The twist is that this search task has to be solved without any hints towards the randomized target location. It may perhaps be just as hard as finding arbitrarily placed glasses in a dark and unknown room for any shortsighted human. Even if the task appears unlearnable at first glance due to the inherent randomness, it is still possible to develop a learnable search pattern, similar to how humans would intuitively approach the problem in a given setting.
Our research tackles the problem of developing an optimal automated approach to explore unknown territory. Experts have manually defined ways to exhaustively traverse a predefined spatial area in an efficient manner such that the area is explored completely. Such search patterns are often used in traditional search tasks with AUVs and originate from considerations of path planning for robotics \cite{GalceranPathplanning}.
Artificial intelligence (AI) has proven to solve complex challenges, and in certain fields is even able to surpass human performance. Therefore, the question arises whether an AI-based approach will outperform expert-defined search patterns. Specifically, reinforcement learning (RL) is well-suited for this task due to its ability to learn optimal strategies through trial and error based on feedback from the environment \cite{sutton2018reinforcement}. 
 
Existing literature on area coverage and efficient exploration \cite{areaCoverage,efficientExploration} focuses on exhaustive area exploration, which differs from our goal of rapidly locating an unknown target - faster than regular exhaustive search patterns. 
Other works have also tackled sparse environments, but their approaches have limited effectiveness on this particular environment, which is sparse, nonstationary, and randomized for every episode. For instance, hindsight experience replay \cite{Hindsightrelabel} is less effective in our particular context, where the target is either found or not, without the use of any intermediate subgoals. 

Our contributions to overcoming the challenges of a sparse, random, and nonstationary environment are as follows:
\begin{enumerate}
    \item Empirical demonstration of the limited effectiveness of tabular Q-learning in a sparse, random, and nonstationary environment.
    \item Developing a Hierarchical Monte Carlo (HMC) approach for accelerating the exploration and moving more efficiently in the environment. The latest reviews in Hierarchical Reinforcement Learning (HRL), such as \cite{HRL_review} and \cite{HRL_survey_2}, indicate a lack of hierarchical approaches for MC. 
    \item Introduction and implementation of the Memory as external Output Filter (MOF), which filters the agent's output (Q-values) to prioritize new state visits and prevent revisits, all while maintaining a manageable state space. 
    \item Application of the fine-tuned HMC approach to our problem, demonstrating superior performance compared to conventional exhaustive search patterns. 
\end{enumerate}

\section{Method}
\label{sec:Method}
This work utilizes reinforcement learning, more precisely, versions of Q-learning and Monte Carlo algorithms, to find an overall strategy to locate randomly generated pollution clouds in the given environment. The following sections demonstrate the reason for the limited capability of tabular Q-learning in the given problem and introduce modifications that turn it into a competitive exploration algorithm. 

At its core, reinforcement learning is based on the Markov Decision Process (MDP), where an agent interacts with an environment, making a sequence of decisions and learning from the consequences. These learned decisions revolve around states, actions, and rewards. The states represent different situations or configurations of the environment as observed by the agent. Given a state, the agent tries to learn the most effective sequence of actions to maximize the cumulative reward it receives over time. Actions are the choices the agent can take to transition from one state to another. The agent's goal is to learn an optimal policy $\pi^*$ as actions that maximize the expected cumulative reward over time. To do this, actions that lead to higher rewards are prioritized, and actions that lead to lower rewards or even penalties are avoided. 

\subsection{Environment}
The environment in this work is modeled as a two-dimensional grid representing an underwater region contaminated with pollution. This grid, along with its discrete action and state space, serves as the foundational setup for applying a model-free reinforcement learning algorithm.

Key components of this environment include the grid itself, the pollution cloud, the initial agent location, the states, and rewards. The grid has a predefined size, where the length determines the number of discrete grid fields, and the movements occur discretely within its boundaries. Each cell can either be unpolluted or contain varying levels of pollution.

The pollution cloud spans multiple grid cells and is characterized by a defined diameter. Its location is generated randomly at the start of each episode, with no prior information given to the agent, reflecting the corresponding real-world oriented setting. This randomness ensures a nonstationary environment, challenging the agent to learn an efficient search pattern.

The AUV typically starts at a predefined initial location, situated in the top-left corner of the grid. An AUV's sensor readings are then mimicked by the environment, outputting feedback in the form of pollution intensity in the current location. These sensor readings are transformed into a normalized reward value for the agent, encouraging it to explore and locate the pollution cloud effectively. 

The agent's actions include moving left, right, up, or down within the grid. The state space is defined by the agent's current position and the pollution intensity at that location. Together, these elements create a dynamic environment with realistic conditions where the agent learns to identify pollution clouds through reinforced exploration.

\begin{figure*}[ht]
    \begin{center}
        \includegraphics[scale=0.46]{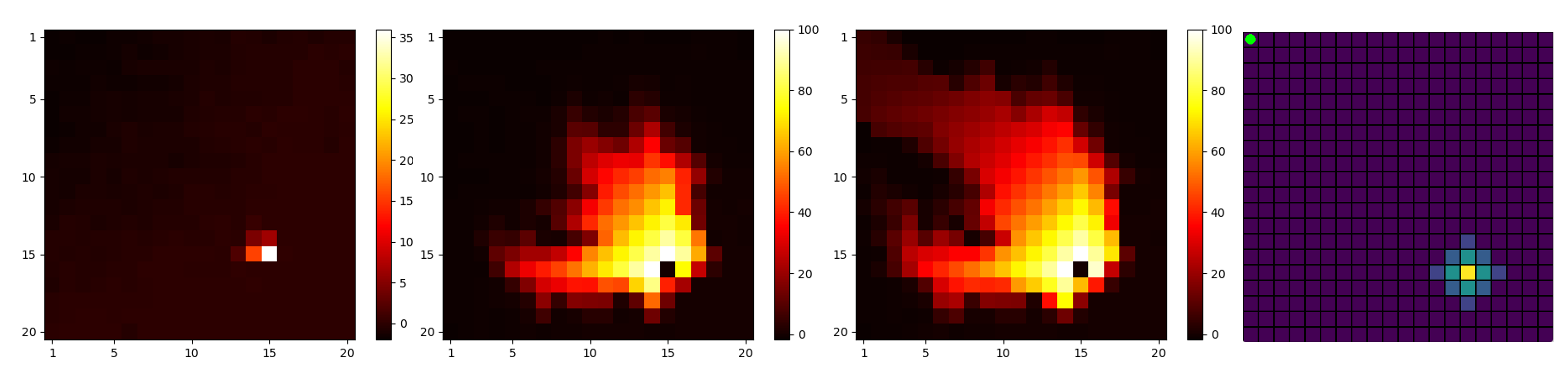}
    \end{center}
    \caption{Q-table visualization for static setting: Showing the maximum Q-value of every state for episodes 500, 1000, 2000 (left to right). The last subfigure shows a visualization of the real environment for comparison.}
    \label{fig:1}
\end{figure*}

\subsection{Limits of Tabular Q-Learning}

To fully understand and highlight the challenge addressed in this work, it is essential to first analyze the limitations of classical approaches. To demonstrate this empirically, we apply tabular Q-learning to a static pollution cloud setting. 

Tabular Q-learning is a model-free RL algorithm, employed to address a wide array of decision-making challenges~\cite{watkins1989learning}. In Q-learning, the agent learns and stores the Q-values $Q$ in a data structure known as the Q-table. These Q-values represent the expected cumulative rewards an agent can achieve by taking a specific action $a$ in a given state $s$.
In \eqref{eq:eq1}, the agents iteratively update their Q-values with the actual reward $R$ and the maximum Q-value by selecting the best action $a'$ in the subsequent state $s'$. Over successive episodes, Q-values converge to represent the expected cumulative reward by taking a specific action and following a particular policy thereafter. 

\begin{equation}
    Q(s, a) \leftarrow Q(s, a) + \alpha \left(R(s,a) + \gamma \max_{a'} Q(s', a') - Q(s, a)\right)
    \label{eq:eq1}
\end{equation}

To manage the exploration-exploitation trade-off, we employed a $\epsilon$-soft-strategy with a linear decay. The state is defined by the agent’s \(x\) and \(y\) coordinates on the grid, with no memory of previously visited locations to prevent an exponential increase in the size of the state space, which would be unmanageable for a tabular Q-table.

Firstly, we consider a static setting where the pollution cloud’s location is randomly generated initially and remains fixed across all learning episodes. The reward is based on pollution intensity measured by the sensors, with an additional reward of 100 for locating the cloud. Fig. \ref{fig:1} illustrates on its right side an exemplary generated cloud within the two-dimensional grid. This Figure also includes three heat maps displaying the highest Q-values for each state in the grid over various episodes to visualize the Q-learning process. 

Initially, all Q-values are initialized to zero. By the 500th episode (first heatmap), the pollution source is occasionally detected through random exploration, resulting in high Q-values near the source, whereas the rest of the Q-table remains unchanged. In episode 1000 (second heatmap), the balance between exploration and exploitation is $\epsilon = 0.5$, indicating an equal likelihood of selecting random or learned actions. Consequently, the source of pollution is found more frequently, leading to increased Q-values around the source, which are getting propagated through the grid world due to the maximum Q-values of the next state accordingly to \eqref{eq:eq1}. By episode 2000 (third heatmap), a clear path from the agent’s starting location to the stationary target is evident, represented by the learned Q-values.

\begin{figure*}[ht]
    \begin{center}
        \includegraphics[scale=0.46]{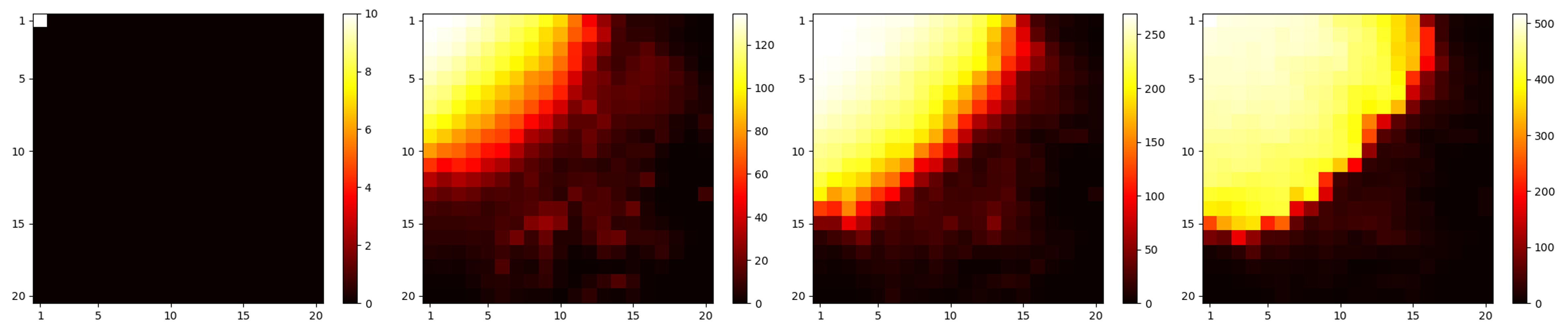}
    \end{center}
    \caption{Q-table visualization for varying settings: Displaying the maximum Q-value of every state for episodes 1, 500, 1000, 2000 (left to right).}
    \label{fig:2}
\end{figure*}

Next, we explore the dynamic setting where the cloud’s location changes with each episode, reflecting real-world task variability. Here, the agent fails to learn effectively due to the continuously and randomly changing target. This can also be visualized by plotting the maximum Q-values of every state in Fig. \ref{fig:2} for different episodes. In the initial learning episode, most maximum Q-values in each state are zero. As learning progresses, high Q-values are primarily observed around the AUV's starting position, whereas smaller Q-values prevail across the rest of the grid.

From these visualizations, two primary conclusions about tabular Q-learning emerge. First, the agent requires numerous episodes in a sparse reward environment to propagate Q-values and discern an optimal path. Second, the agent learns the position of the cloud, which is constantly changing in the environment, in each episode. Even if the agent were able to propagate all Q-values based on the final reward correctly through the whole grid in one episode and learn the final position as shown in the third heat-map of Fig. \ref{fig:1}, in the next episode, the cloud would be randomly generated in a different spot, and all information gained would be obsolete. 

Thus, simply learning the optimal path based on the target position with Q-learning is insufficient, and the goal is to learn the optimal path itself directly without the need to propagate the Q-values over numerous episodes. Therefore, we introduce the following modifications to Q-learning to effectively address sparsity and nonstationary randomness, which results in a competitive exploration algorithm.

\subsection{Q-learning Modifications}
\label{sec:qManipulations}
In the original task description, the goal is to consistently find the unknown position of a pollution cloud in a relatively vast and empty environment. 
Tabular Q-learning is not able to achieve that, as gained rewards represent the cloud position in one episode, which does not reflect the cloud position in the next episode due to the random and nonstationary environment. This randomness makes it infeasible to construct a consistent path using Q-table updates based solely on rewards linked to the current location of the cloud.
Therefore, the algorithm needs to overcome the randomness of the cloud’s location in the grid world to learn an overall strategy. However, one can make adjustments to this learning process to mitigate the given problems and enhance the capability of the Q-table to represent the best strategy in such sparse and dynamic environments.

\subsubsection{Hierarchical Reinforcement Learning} 
Within the given 2D grid, the agent's need to explore competes with the need for fast traversal of the grid. This results in an agent configuration struggling, or even failing to develop a consistent and stable movement behaviour. Grouping several action decisions into one decision can cover more area of the state space in fewer decisions made, effectively enabling the agent to cover a larger area of the grid with less random movement, stabilizing the exploration \cite{HRL_review, nachum2019does}. 

Given that the pollution cloud is assumed to be larger than a single grid cell, it follows that not every grid cell needs to be explored. In our approach, we implement action groupings of sets of unidirectional steps, referred to as options, as introduced by \cite{SUTTON1999181}. With $o_i$ being the option to go in direction $i$ (as before, either up, down, left, or right), it can be represented by consecutive actions $a_i$ as shown in \eqref{eq:option}. 
\begin{equation}
    o_i = (a_i,a_i,a_i,...) \in \mathbb{N}^J
    \label{eq:option}
\end{equation}

Here, $J$ specifies the number of steps taken within an option, known as the option length. The optimal option length theoretically depends on the size of the environment and the size of the pollution cloud. Consequently, within the scope of this study, the option length will be fine-tuned to determine the optimal value for our environment configuration. In real-world settings, the knowledge gained from this tuning can be exploited, as both environment size and pollution cloud size can be inferred. Typically, a search area is defined prior to deploying an Autonomous Underwater Vehicle (AUV), and an assumption about the size of pollution clouds can be made.

\subsubsection{Multiple Goal Learning} 

\begin{figure*}[ht]
    \begin{center}
        \includegraphics[scale=0.24]{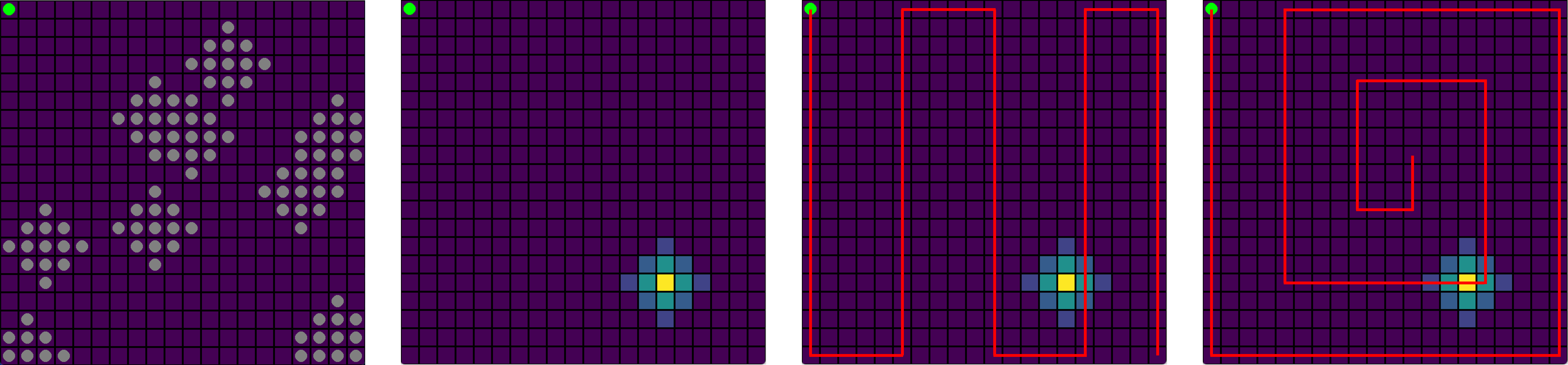}
    \end{center}
    \caption{From left to right: Training environment with a number of randomly spawned clouds. Evaluation environment with one randomly spawned cloud. Evaluation pattern, Snake and Spiral, respectively, in the evaluation environment.}
    \label{fig:pattern}
\end{figure*}

Whereas hierarchical approaches enhance the agent's ability to traverse the environment effectively, they do not address the inherent reward sparsity. In the given environment, only the target (the pollution cloud) provides an explicit reward, and reaching it represents solving the exploration task. 
The proposed solution for this is a change in training conditions, so that the agent is tasked to search for a multitude of randomly located clouds within one epoch, as in Fig. \ref{fig:pattern} on the left side. The agent receives a reward depending on the number of clouds found in as few steps as possible.
While the ultimate task of finding one pollution cloud in an environment remains the same (second grid of Fig. \ref{fig:pattern}), the agent can now train to do this on many differently located clouds at once, therefore building knowledge about how to search for a cloud. In other words, a step taken by the agent needs to be good for multiple locations of clouds and not only for one single location.

\subsubsection{Trajectory Reward Learning} 
\label{sssec:TRL}
In the process of learning the long and complex trajectory required to reach a pollution cloud, the agent receives a reward only upon reaching the cloud.
However, the intermediate steps, although crucial, do not yield any immediate rewards. To account for the importance of these steps, we propose updating all Q-values of a successful trajectory based on the average reward of that trajectory. 
This approach allows the algorithm to learn possible trajectories more effectively, as it updates all steps collectively after completing the search process rather than iteratively step by step.
This reward shaping approach is closely related to the classic Monte Carlo method for learning Q-values, where the whole trajectory is being unrolled and the accumulated reward of the complete route is utilized for updating \cite{sutton2018reinforcement}. 
By combining the multiple goal learning and all step learning approaches, the reward function needs to be reshaped:
\begin{equation}
    R_{T} = \frac{S_r}{n_{step}} {n_{poll}}
    \label{eq:all_step_reward}
\end{equation}
In \eqref{eq:all_step_reward}, the reward $R_{T}$ is once generally calculated for every single step of the trajectory $T$ by using the total number of pollution clouds $n_{poll}$ found in one learning episode. Additionally, the reward is normalized by the total number of steps $n_{step}$ needed to locate all of the generated clouds in the episode. This shaping of reward induces the agent to locate the clouds as fast as possible, which is one of the main purposes of the given task. Finally, the reward is scaled by the factor $S_r$, which serves as a hyper-parameter to optimize the learning procedure (see Appendix \ref{app:1}). 

\subsubsection{Memory As Output Filter}
Incorporating a memory that records all or part of previously visited locations is impractical due to an exploding state space. Such state-space growth is problematic given the limited abstraction ability of the tabular Q-learning approach compared e.g. to deep Q-learning. However, retaining information about previously explored states within an ongoing episode can be significantly advantageous for searching in a grid with numerous locations.
Inserting a penalty for revisiting locations into the training process without having any information about already explored fields as input for the agent harms the learning enormously. This is a consequence of the fact that the agent would be able to receive two different rewards for the exact same input, and consequently tries to learn two different and conflicting Q-values for the same state. Thus, the memory needs to influence the agent in an alternative manner.
To achieve this external addition of memorized information, we included the memory of former visited states as an output filter to the agent, filtering out states that have been visited before. 
More precisely, when the agent evaluates possible actions based on Q-values in the current state, the memory component (0 if not visited, 1 if visited) is externally subtracted from the corresponding Q-value, before a decision is made. This suggests the agent select the next highest Q-value if the preferred state has been visited before. A subsequently introduced MOF value scales the magnitude of the memory, influencing how strongly the memory influences the decision-making.

In \eqref{eq:MOF}, the memory component $M$ (indicating the number of visits to potential next states $s'$) and the scaling factor $S_{MOF}$ are subtracted from the Q-values of the current state-option-pairs, which would lead to the next possible states $s'$. 
Already visited states and border states in the memory array $M$ are increased by 1 for each visit (the memory array is initialized with zero values). By reducing the Q-value retrospectively with the MOF, he agent is discouraged from revisiting states. The effect of this reduction depends on the scaling factor $S_{MOF}$, which requires fine-tuning. 
During decision-making, the agent selects the best option $o^*$ by maximizing the adjusted Q-values. It's important to note that the MOF requires distinct treatment for exploration and exploitation within the $\epsilon$-soft strategy since exploration should be random. This can be achieved by eliminating the impact of the memory filter during exploration ($S_{MOF} = 0$). 

\begin{equation}
    o^* = \argmax_{o} \bigg(Q(s, o) - S_{MOF} * M(s')\bigg)
    \label{eq:MOF}
\end{equation}
The inserted memory element is explicitly not part of the learning and updating process, and it also does not change the Q-values inside the Q-table in any way. By subtracting the memory component rather than overwriting the Q-values, this method proves especially beneficial when all potential next states have been explored, ensuring the agent still chooses the best available option based on the highest probability.

\subsection{Evaluation} 
\label{seq:Eval}

\begin{figure*}[ht]
    \begin{center}
        \includegraphics[scale=0.54]{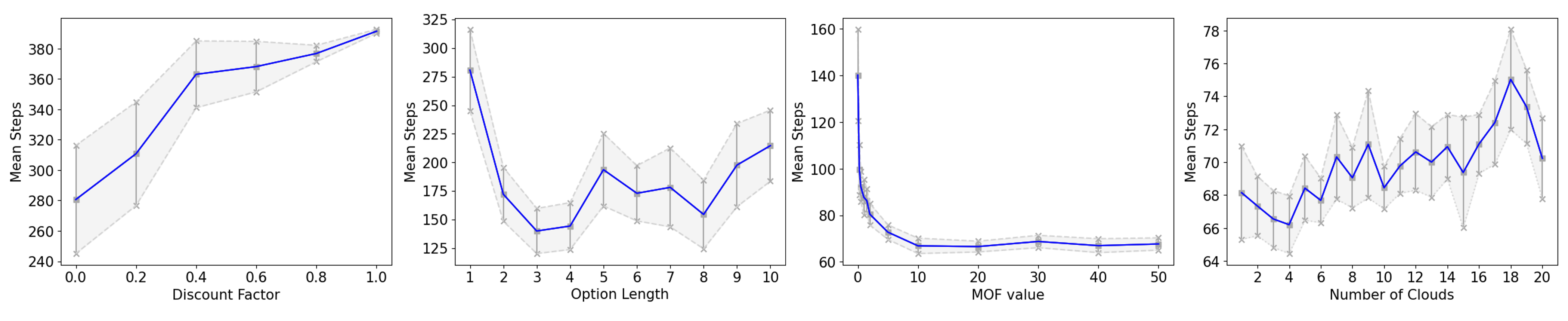}
    \end{center}
    \caption{Results of the first parameter tuning loop: Each graph shows the mean performance across 1000 evaluation episodes, each associated with one of the 20 independent runs of our method. The graphs include a 95\% confidence interval, shown as the greyed out area around the blue curves. The y-axis presents the performance (mean number of steps until the agent discovers the pollution cloud). From left to right: tuning of the discount factor $\gamma$, option length, MOF value, and number of clouds during training, respectively.}
    \label{fig:tune}
\end{figure*}

The goal of applying reinforcement learning to this sparse, nonstationary, and randomized exploration task is to discover efficient exploration and area coverage strategies that outperform simple exhaustive search methods, such as simply visiting every single location. 
Ideally, it should also overcome the performance of exploration strategies designed by experts in AUV control for discovering sources of pollution. 
Given the grid size and a reasonable assumption about pollution cloud size, it is feasible to design more efficient strategies than visiting every single grid location, where the traversed path is sufficiently tight to ensure no pollution cloud fits in the gaps.
To evaluate the RL-agent, we compare it against two manually designed patterns: "Snake" and "Spiral." Both of these patterns can be seen as derivatives of what is referred to as a seed spreader motion by \cite{GalceranPathplanning}.
The path these patterns utilize can be observed on the right side of Fig. \ref{fig:pattern} for a given grid length of 20 and a pollution diameter of 5. The patterns were constructed such that they may be adapted depending on these two parameters. 
Within the evaluation phase, these two patterns, as well as the trained RL agent, operate in the same environment for 1000 iterations. 
In each iteration, they search for one unknown pollution cloud positioned randomly within the grid, as shown in the second grid of Fig. \ref{fig:pattern}. The number of steps required to locate the pollution cloud is recorded for each iteration.

After completing all the evaluation iterations, the performance of the three competitors is assessed based on the following metrics:

\begin{itemize}
    \item \textbf{Average Steps}: The mean number of steps required by each approach is calculated for the 1000 iterations with random cloud locations. This metric allows us to compare the overall efficiency of the trained agent with the predefined patterns.
    \item \textbf{Score of 1000 duels}: A score is computed to indicate how many of the 1000 iterations with random cloud location the RL-agent completed faster than the Snake and Spiral patterns. This metric assesses the performance irrespective of average speed, countering any potential distortions caused by speed variations.
    \item \textbf{Score-Mapping}: A plot is created to show the wins, ties, and losses of the RL-agent versus the two patterns for every possible cloud location in the grid (400 locations). This visualization in the scoring map helps identify specific areas where the RL-agent outperforms the exhaustive search pattern.
\end{itemize}
These metrics collectively provide a comprehensive assessment of the RL-agent's performance compared to traditional search patterns.

\section{Results}
\label{sec:Results}

Due to the high number of testable hyperparameters and algorithmic design decisions (discount factor, option length, MOF value, number of clouds during training), an exhaustive search is deemed unfeasible. Therefore, the RL agent is optimized using an iterative manual hyperparameter optimization procedure. One after another, the modifications described in Section \ref{sec:qManipulations} are introduced to the Q-learning algorithm and optimized with respect to the average performance of 20 independent runs (e.g., executing the complete algorithm-learning and evaluating 20 times). Specifically, one parameter is tuned at a time; the best-performing value of that parameter is then fixed while tuning the subsequent hyperparameters, continuing until all hyperparameters are tuned.

The evaluation of the final optimized RL agent is done by comparing the average number of steps required to find the clouds in 1000 independent evaluation episodes, using the method described in Section \ref{seq:Eval}. This iterative process is executed twice: The first loop finds the best value for each tuned parameter, and the second confirms the findings from the first loop and refines any parameters that might have been affected by subsequent tuning. Environmental settings (grid size, pollution size) and certain hyperparameters (maximum steps per episode, $\epsilon$, and its decay) are predefined and kept constant during optimization. The parameters with the highest impact on evaluation performance are plotted in Fig. \ref{fig:tune}.

\begin{figure*}[ht]
    \begin{center}
        \includegraphics[scale=0.49]{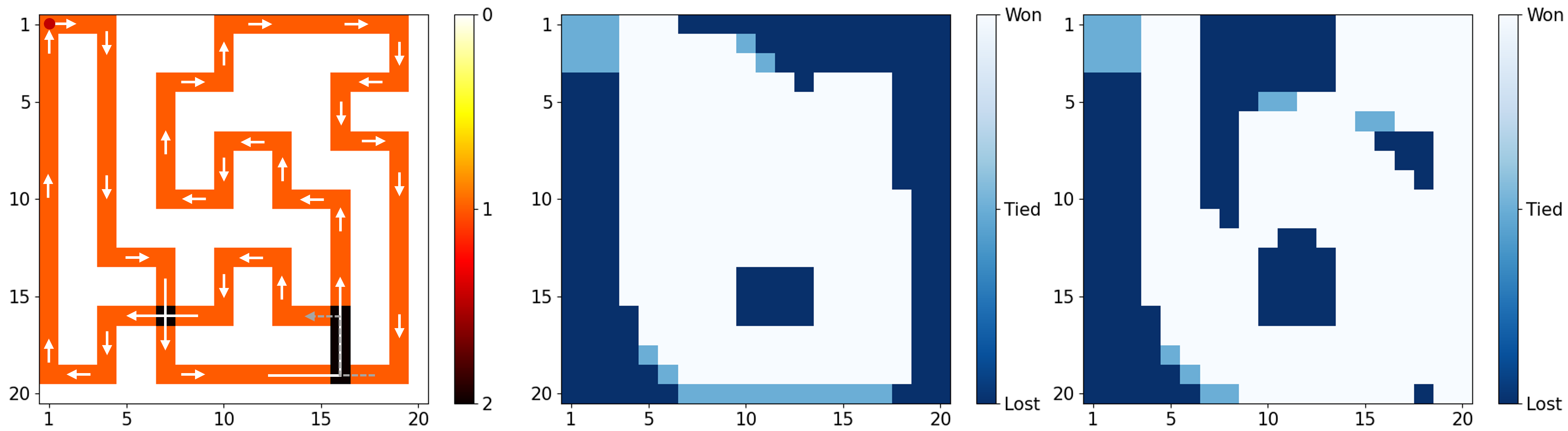}
    \end{center}
    \caption{The route followed by our agent, visualized with a heat-map coloring representing the number of visits per state (left). Duels won against the Spiral (middle) and against the Snake (right) by generating a cloud with the center in every location of the grid.}
    \label{fig:route}
\end{figure*}

\textbf{Q-learning and Discount Factor Tuning:}
The first tests focus on the magnitude of the discount factor ($\gamma$), revealing a mean number of steps (between 392 and 395), comparable to the maximal episode length (400). Subsequently, Trajectory Reward Learning (Section \ref{sssec:TRL}) is implemented using the reward defined by \eqref{eq:all_step_reward}, which reduces the number of steps due to the decreasing discount factor $\gamma$ (Fig. \ref{fig:tune}, left). A discount factor of 0 results in higher performance, transforming the update rule from the Q-learning setting as in \eqref{eq:eq1} into a Monte Carlo-based approach as in\eqref{eq:MC}.

\begin{equation}
    Q(s, a) \leftarrow Q(s, a) + \alpha \left(R_T - Q(s, a)\right)
    \label{eq:MC}
\end{equation}

\textbf{Option Length:}
Various option lengths are evaluated, showing, as expected, dependency on a combination of grid and pollution cloud size. The optimal option length is found to be 3.

\textbf{MOF Value:}
The optimization curve for the MOF value flattens out at 10, indicating that higher MOF values filter out revisiting states similarly, and more strictly, than lower values corresponding to \eqref{eq:MOF}.

\textbf{Number of Clouds:}
Testing different numbers of clouds during training indicates that fewer clouds result in better-performing policies, with a nonsignificant trend favoring four clouds. Generating more clouds increases the steps required to find all clouds during a training episode, as it takes longer to locate all of them.

Other tested parameters do not produce significantly better results than the baselines and are therefore discussed in Appendix \ref{app:1}. Traditional RL hyper-parameters (learning rate and episode number) are also optimized, as detailed in Table \ref{tab:bestParameters} in Appendix \ref{app:2}, which also shows the predefined hyper-parameters. 

In the second loop of iterative optimization, hyperparameters largely remain the same as their initial optimization. Performance is evaluated not only on the average number of steps to identify the pollution cloud but also on the number of duels won against competing search patterns. As described in Section \ref{seq:Eval}, the higher the winning score against these patterns, the better the performance of the agent, indicating a better-tuned parameter value. The significant adjustment in the second loop is the number of clouds, with the best performance using one or two clouds. Moreover, it is noteworthy that the agent performed on average better than the Snake and Spiral already after the first tuning loop and improved slightly in the second tuning loop.

The left plot in Fig. \ref{fig:route} shows the search path learned by the RL agent, capable of beating Snake and Spiral. It shows that faster movement through the central region of the grid is prioritized to traverse a larger area in fewer steps. This leads to the following results in every explained evaluation metric:
\begin{itemize}
    \item \textbf{Average Steps}: The fine-tuned RL-agent shows a median (mean) step number of 43 (53.49) compared to 54 (53.51) and 73 (66.74) for the expert-designed search patterns, Snake and Spiral, respectively. (Distribution graphs of the agents' performance are displayed in Appendix \ref{app:2})
    \item \textbf{Score of 1000 duels}: In the 1000 randomized and nonstationary evaluation iterations, the RL agent wins (ties) 643 (47) times against Snake and 583 (62) times against Spiral.
    \item \textbf{Score-Mapping}: By evaluating every one of the possible 400 grid locations, the model wins (ties) 257 (17) times against Snake and 230 (23) times against Spiral. The two right plots of Fig. \ref{fig:route} show the results of the 400 duels against Snake and Spiral. Wins are mostly achieved in the central area of the grid, indicating the successful learning of a useful search pattern heuristic.
\end{itemize}

\section{Discussion}
\label{sec:Discussion}

In a reward-sparse environment with randomized and nonstationary targets for each episode, it is challenging for tabular Q-learning to develop effective search strategies. 
However, a Monte Carlo-based RL algorithm (Trajectory Reward Learning with $\gamma=0$), when supplemented with certain modifications, can effectively learn to navigate such environments. 
Our study demonstrates that even with optimized grid traversal patterns based on assumptions of minimal pollution cloud size, the modified Monte Carlo algorithm outperforms these expert-designed search patterns and tabular Q-learning in terms of average performance.

These results are promising not only for AUV exploration but also as a potential solution to the broader problem of navigating sparse and nonstationary environments with randomly placed targets. 
Both hierarchical reinforcement learning and the use of Memory as Output Filter (MOF) are crucial components of this solution. The commitment to multiple steps in the same direction through options helps reduce jittery movements, facilitating the learning of effective trajectory patterns. Similarly, MOF enhances the learning process by integrating memory without altering saved Q-values, making it an efficient way to incorporate memory without increasing the state space significantly. However, it is noteworthy that strictly speaking, the Markov property is lost with such an approach.

Evaluating the modified RL approach against the baseline search patterns proves to be a sensible method of performance assessment, as these patterns are well-adapted versions of exhaustive search strategies that leverage knowledge of grid and cloud size. 
Additionally, although learning multiple goals simultaneously initially appeared promising, it did not significantly impact performance in the final parameter tuning loop. Given the increased training time, utilizing a higher number of goals is often not advisable.

Nevertheless, this study has several limitations. Whereas the current setting approximates a potential AUV environment, it does not fully reflect real-world conditions. Further investigations can be pursued in two main directions. 

First, insights from this study could be applied to deep reinforcement learning, which can handle larger state spaces. Training could include a more dynamic and realistic simulated environment with, e.g. an adaptable grid, varying cloud sizes, a continuous state space, or simulated underwater currents. Additionally, future projects could include a more extensive option search with different semantics or converting the MOF into an intelligent filter by using a neural network. 

Second, this methodology may be applicable to other scenarios, such as navigating sparse or partially observable environments with unknown or randomized goal locations. Sparse environments are common in oceans but can also be found in the air or on land, e.g., under adverse weather conditions.

\bibliography{Sources}

\newpage

\newpage
\appendices

\section{Further Q-Learning Modifications and Hyper-Parameter Tuning}
\label{app:1}

Some of the tested modifications did not exhibit the expected effects during parameter testing. Low-impact, nonsignificant, or even negative results are still valuable to the scientific process. Therefore, these results are included in the Appendix with a brief analysis to further our understanding. All modifications were evaluated using the iterative process described in Section \ref{sec:Results} and the performance metrics explained in Section \ref{seq:Eval}.

\subsection{Best Experience Learning}

After testing various modifications, we found that the random exploration can impede effective learning. Specifically, with our approach of "Trajectory Reward Learning," every state-action transition (or state-option transition in HRL) and its Q-value along the trajectory is assigned an equal reward, normalized by the number of steps required to locate the pollution cloud. Consequently, a state-action transition may appear efficient in one episode but inefficient in another. This leads to conflicting rewards and divergent Q-values for the same state-action transition, impeding consistent learning. This issue is exacerbated by the sparse, random, and nonstationary nature of our environment.

\begin{figure*}[!h]
    \begin{center}
        \includegraphics[scale=0.34]{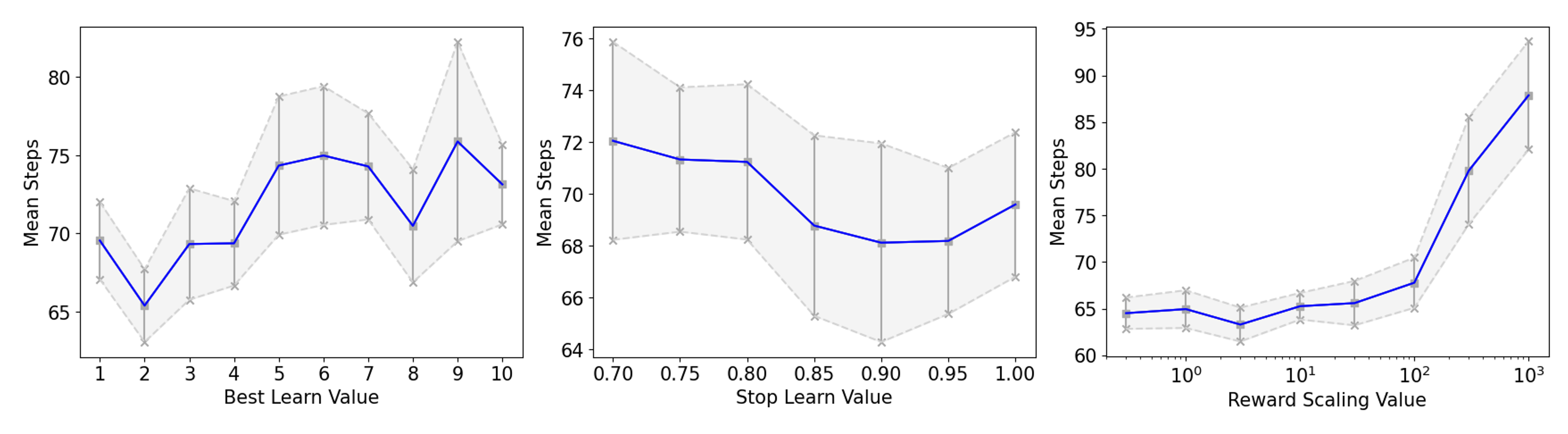}
    \end{center}
    \caption{Results of the parameter tuning: Each graph shows the mean performance across 1000 evaluation episodes, each associated with one of 20 independent runs of our method. Each graph includes a 95\% confidence interval, shown as the greyed-out area around the blue curves. The y-axis represents performance (mean number of steps until the agent discovers the pollution cloud). From left to right: Best Learn Value, Stop Learn Value, and Reward Scaling Value (with a logarithmic x-axis).}
    \label{fig:app1}
\end{figure*}

Inspired by "survival of the fittest," a new approach filters out mediocre training epochs by only inserting the best-performing experiences from a set of training epochs into the learning process. This means the distribution of clouds and their locations remains constant for a predefined number of learning runs (Best Learn Value), and only the attempt with the highest reward is utilized for training in this episode. The highest reward is typically received for finding the most clouds and/or using the fewest steps, in accordance with the reward shaping in \eqref{eq:all_step_reward}. The left side of Fig. \ref{fig:app1} shows the results calculated as described in Section \ref{sec:Results}. The graph indicates a trend toward higher performance with lower Best Learn Values. Despite slightly better results for a Best Learn Value of two, we chose to continue with the value of one due to the increased training time resulting from this modification.

\subsection{Further Tuning Results}
In addition to major modifications, we also fine-tuned other hyperparameters. The center graph in Fig. \ref{fig:app1} analyzes the impact of stopping the learning process earlier (Stop Learn Value). The Stop Learn Value represents the relative number of episodes completed before learning was terminated. This experiment was introduced because some better results were occasionally found in earlier stages of the training process. However, this trend did not carry over to the optimized version of the agent.

The right subplot of Fig. \ref{fig:app1} shows the evaluation results for the Reward Scaling Value, which scales the reward on a logarithmic scale to examine its impact on training and evaluation results. The trend indicates distinctly worse performance for a Reward Scaling Value above 100, whereas below 30, the mean steps to find the cloud remains consistently good within the scope of measurement accuracy. 

\begin{figure*}[!h]
    \begin{center}
        \includegraphics[scale=0.34]{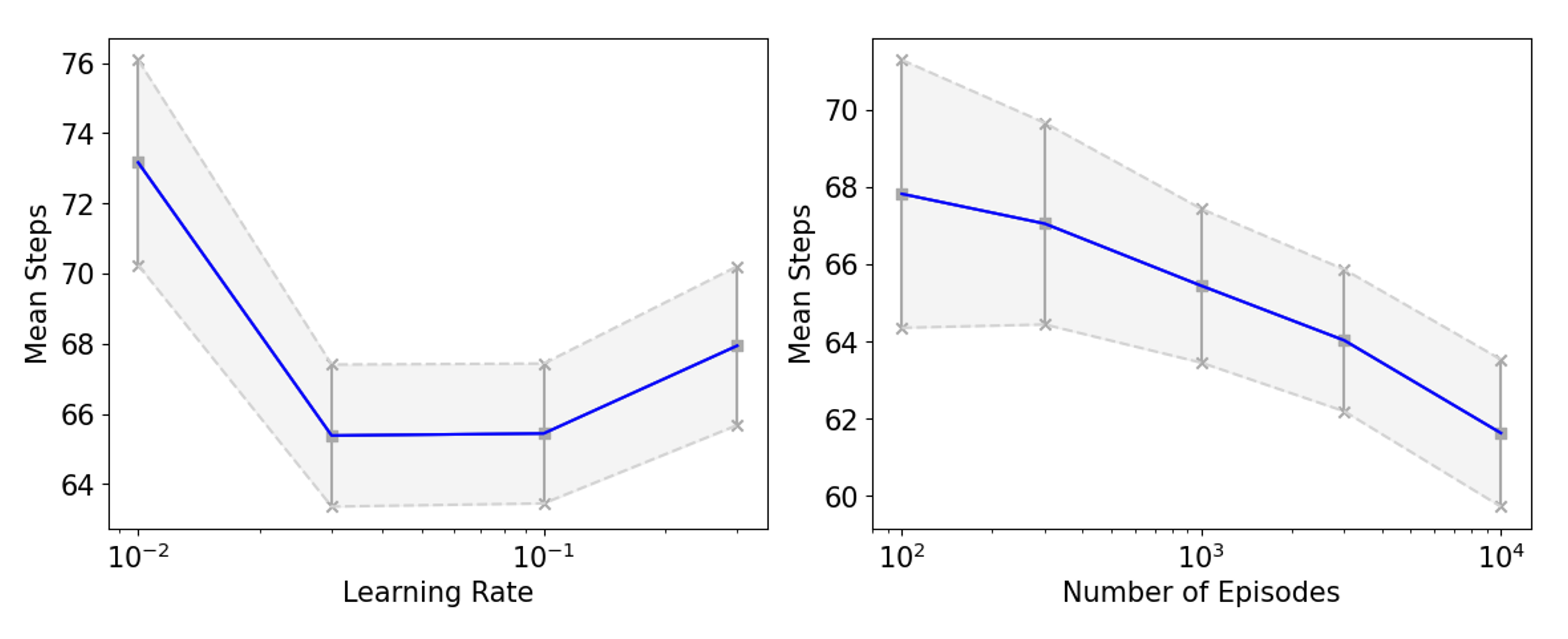}
    \end{center}
    \caption{Results of the parameter tuning: Each graph shows the mean performance across 1000 evaluation episodes, each associated with one of 20 independent runs of our method. Each graph includes a 95\% confidence interval, shown as the greyed-out area around the blue curves. Both Learning Rate (left) and Number of Episodes (right) use a logarithmic scale for the x-axis.}
    \label{fig:app2}
\end{figure*}

The parameter-tuning results for the learning rate of our agent are presented in Fig. \ref{fig:app2} (left). The learning rate directly influences the balance between exploration and exploitation during the learning process, determining how quickly the agent adapts its knowledge. By adjusting the learning rate, we aim to optimize the agent's ability to gradually converge towards optimal policies, while still allowing sufficient exploration of the state-action space. This fine-tuning ensures the agent efficiently learns from its experiences and improves its decision-making over time. Based on these results, we set the learning rate to 0.1.

Fine-tuning the number of learning episodes is essential, as it directly impacts the agent's ability to converge to an optimal policy. Adjusting this parameter helps balance thorough exploration and avoiding excessive training, which can lead to overfitting or inefficient resource utilization. Ultimately, optimizing the number of learning episodes ensures effective learning within a reasonable timeframe and achieves the desired performance level. Additionally, the linear $\epsilon$-decay was normalized with the number of episodes to spread exploration and exploitation evenly throughout the training process. Whereas the right graph of Fig. \ref{fig:app2} shows improved performance with more episodes, the numerical enhancement is marginal. Given the substantial increase in training time, this slight improvement does not justify higher episode numbers for the parameter fine-tuning loops.

\section{Optimal Parameter Configuration}
\label{app:2}
\begin{figure*}[!h]
    \begin{center}
        \includegraphics[scale=0.3]{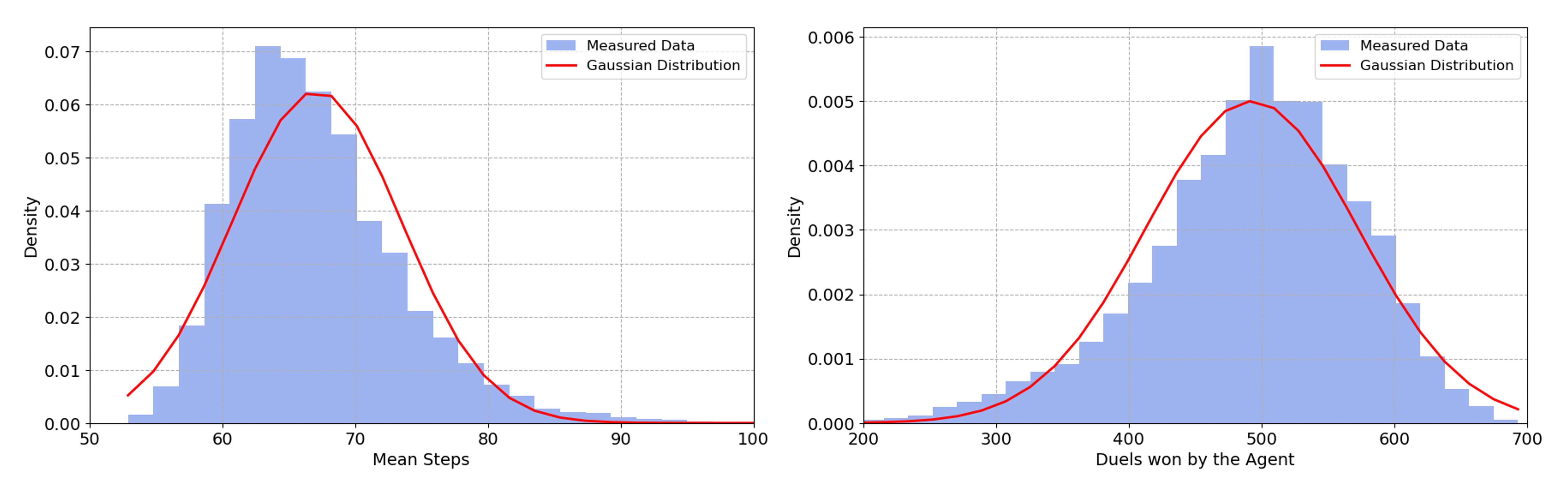}
    \end{center}
    \caption{Performance distributions for 10,000 trained agents. The left plot shows the average number of steps to find the cloud in 1,000 evaluation episodes. The right plot shows the percentage of victories against the Snake, measured by how often the agent locates the cloud in 1,000 evaluation episodes.}
    \label{fig:app3}
\end{figure*}
This section reports the parameters and results relevant to the optimal settings. Table \ref{tab:bestParameters} shows the final configuration of both predefined and fine-tuned parameters. These optimal settings were used to run the algorithm 10,000 times, training and evaluating 10,000 different agents. This extensive approach helps us identify the best-performing agents while also creating a performance distribution. This distribution is important as the results and discovered policies vary due to the random nature of exploration and the environment.

\begin{table}[htbp]
    \caption{Hyper-parameter settings for the best performing agent. The first three values are predefined environmental parameters, whereas the rest are optimized hyperparameters based on our tuning procedures.}
    \begin{center}
        \begin{tabular}{ll}
            \multicolumn{1}{l}{\bf PARAMETER NAME}  &\multicolumn{1}{l}{\bf VALUE}
            \\ \hline \\
            Grid Length                         &20 \\
            Pollution Diameter                  &5 \\
            Max Steps per Episode               &400 \\
            Number of Episodes                  &1000 \\
            Learning Rate                       &0.1 \\
            Discount Rate ($\gamma$)            &0 \\
            $\epsilon$ (Start, Final, Decay)    &1.0, 0, 0.001 \\
            Best Learn Value                    &1 \\
            Number of Clouds                    &1 \\
            MOF Value                           &10 \\
            Stop Learn Value                    &1 \\
            Option Length                       &3 \\
            Reward Scaling Value                &30 \\
        \end{tabular}
    \end{center}
    
    \label{tab:bestParameters}
\end{table}

Fig. \ref{fig:app3} illustrates the performance distribution of the 10,000 trained agents using these optimal parameters. The left plot displays the average number of steps taken to find the cloud across 1,000 evaluation episodes for each trained agent. The right plot shows the percentage of victories in duels against the Snake, where the agent locates the cloud within 1,000 evaluation episodes.

The distribution plots reveal that while our agents generally perform well, there is variability due to the random nature of exploration and the environment. Most agents find the cloud efficiently, with some exhibiting exceptional performance and others showing reduced effectiveness.

\end{document}